\documentclass[a4paper,12pt]{elsarticle}

\usepackage{amssymb}

\usepackage{multirow}
\usepackage[table]{xcolor}
\usepackage{float}
\usepackage{placeins}
\restylefloat{table}

\definecolor{Gray}{gray}{0.7}

\journal{Engineering Applications of Artificial Intelligence}

\begin{document}

\begin{frontmatter}



\title{Semantic Segmentation using Vision Transformers: \\A survey}


\author[inst1]{Hans Thisanke}
\ead{e16368@eng.pdn.ac.lk}
\author[inst1]{Chamli Deshan}
\ead{e16076@eng.pdn.ac.lk}
\author[inst1]{Kavindu Chamith}
\ead{e16057@eng.pdn.ac.lk}

\author[inst2,inst3]{Sachith Seneviratne}
\ead{sachith.seneviratne@unimelb.edu.au}
\author[inst2,inst3]{Rajith Vidanaarachchi}
\ead{rajith.vidanaarachchi@unimelb.edu.au}
\author[inst1]{Damayanthi Herath\corref{cor1}}
\ead{damayanthiherath@eng.pdn.ac.lk}

\cortext[cor1]{Corresponding author}

\affiliation[inst1]{organization={Department of Computer Engineering},
            addressline={University of Peradeniya}, 
            city={Peradeniya},
            postcode={20400}, 
            country={Sri Lanka}}

\affiliation[inst2]{organization={Melbourne School of Design},
            addressline={University of Melbourne}, 
            city={Parkville},
            postcode={VIC 3010}, 
            country={Australia}}

\affiliation[inst3]{organization={Faculty of Engineering and IT},
            addressline={University of Melbourne}, 
            city={Parkville},
            postcode={VIC 3010}, 
            country={Australia}}

\begin{abstract}
Semantic segmentation has a broad range of applications in a variety of domains including land coverage analysis, autonomous driving, and medical image analysis. Convolutional neural networks (CNN) and Vision Transformers (ViTs) provide the architecture models for semantic segmentation.  Even though ViTs have proven success in image classification, they cannot be directly applied to dense prediction tasks such as image segmentation and object detection since ViT is not a general purpose backbone due to its patch partitioning scheme. In this survey, we discuss some of the different ViT architectures that can be used for semantic segmentation and how their evolution managed the above-stated challenge. The rise of ViT and its performance with a high success rate motivated the community to slowly replace the traditional convolutional neural networks in various computer vision tasks. This survey aims to review and compare the performances of ViT architectures designed for semantic segmentation using benchmarking datasets. This will be worthwhile for the community to yield knowledge regarding the implementations carried out in semantic segmentation and to discover more efficient methodologies using ViTs.
\end{abstract}




\end{frontmatter}


\section{Introduction}


Transformers became the new state-of-the-art in natural language processing (NLP) \cite{Vaswani2017} after the tremendous success it achieved. This led to the development of ViT \cite{Dosovitskiy2020} which was later adapted into the computer vision tasks such as image classification \cite{Dosovitskiy2020, chen2021crossvit}, semantic segmentation \cite{liu2021swin, zheng2021rethinking} and object detection \cite{zhu2020deformable, dai2021dynamic}. A typical Transformer encoder consists of a multi-head self-attention (MSA) layer, a multi-layer perceptron (MLP), and a layer norm (LN). The main driving force behind the ViT is the multi-head self-attention mechanism. It helps ViT to capture long-range dependencies with less inductive bias \cite{park2022vision}. When trained on a sufficient amount of data, ViT shows remarkable performance, beating the performance of state-of-the-art CNNs \cite{Dosovitskiy2020}. However, ViTs still have some drawbacks compared to CNNs such as the need for very large datasets. Strategies such as self-supervised based approaches can be used to alleviate some of these drawbacks and further enhance ViTs \cite{caron2021emerging}. 

Semantic segmentation is the process of assigning a class label to each and every pixel of an image. This requires accurate predictions at the pixel level. For segmentation, there exist both CNN-based models and Transformer based models. However, plain ViT models cannot be directly used for segmentation tasks because they do not consist of segmentation heads \cite{xu2021efficient}. Instead SETR \cite{zheng2021rethinking} and Swin Transformer \cite{liu2021swin} based architectures can be utilized for segmentation tasks. Unlike image classification, dense prediction tasks such as semantic segmentation and object detection come with a few difficulties due to the rich intra-class variation, context variation, occlusion ambiguities, and low image resolution \cite{strudel2021segmenter}. There have been many improvements in the ViT domain in the last few years to overcome these challenges while further developments are still in progress to make them efficient. 

The review focuses specifically on semantic segmentation using Vision Transformers. The comparison of the ViT models specialized for semantic segmentation is discussed with architecture-wise and tabulated specific sets of model variants that can be compared with the same set of benchmark datasets. The current surveys performed on ViTs have been structured with a detailed historical evolution from NLP to the Vision Transformer domain. \cite{khan2022transformers} focuses on self-attention and its varieties with advantages and limitations with existing methods for segmentation, object detection, classification, and action recognition. The comparison follows between CNN and ViT backbones on the ImageNet dataset. 
The survey done by \cite{liu2023survey} is also considering various vision tasks and surpasses CNN-based models with experimental results on benchmark datasets. Even though several surveys have been done \cite{khan2022transformers,liu2023survey,han2022survey}, a comparison between segmentation models with several benchmark datasets to identify the best-performing model has not been performed. In our survey, we provide a set of segmentation models, for each of which we define the best variant in each benchmark dataset category. This is useful in the sense of identifying the most optimal parameters such as patch size, iterations count for each variant of the model. By providing mIoU (\%) of model performance results over several semantic segmentation-related benchmark datasets, overall evaluation and highest-performing model variants for each dataset can be identified.

In Section \ref{Semantic Segmentation using Vision Transformers} we discuss the applications of semantic segmentation, ViTs, their challenges, and loss functions. Section \ref{Datasets} describes benchmark datasets used in semantic segmentation. Section \ref{Meta - analysis} describes the existing work done in semantic segmentation using ViTs and presents a quantitative analysis. Finally, Section \ref{Discussion} provides the discussions and Section \ref{Conclusions and Future Directions} concludes the paper with future directions. 

\section{Semantic Segmentation using Vision Transformers} \label{Semantic Segmentation using Vision Transformers}

This section aims to provide an in-depth analysis of the applications in semantic segmentation, with a focus on recent advancements in ViTs. We begin by exploring the principles and architecture of ViTs and their potential for improving semantic segmentation performance. We then delve into various application domains of semantic segmentation. We also devote a section to practical approaches for overcoming the data limitations that often arise in ViT models. Finally, we discuss various loss functions used in semantic segmentation and their effectiveness in different scenarios.

\subsection{Vision Transformers}

Automatic segmentation techniques have been evolving and improving throughout the years with the advancements of deep learning approaches and the application of semantic segmentation in practical usage. For semantic segmentation, the requirement is to locally identify the different classes in the image with spatial location. For that, the fully connected layers in the conventional CNN architecture were replaced with fully convolutional layers combined with feature extraction. This was introduced as Fully Convolutional Networks (FCN) \cite{long2015fully} to identify
high-level semantic features from images. These networks have shown to be faster compared to previous CNN-based techniques and are also capable of generating segmentation maps for images of any resolution. Some of the commonly known architectures are U-Net (state-of-the-art FCN) and more improved architectures with higher accuracy and efficiency are developed by \cite{oktay2018attention,diakogiannis2020resunet,zhou2018unet++}.

One of the limitations identified with the FCN architecture is the low resolution of the final output segmentation image of the feature map due to going through several convolutional and pooling layers. Furthermore, the locality property of the FCN-based methods caused limitations to the capture of long-range dependencies of the feature maps. To solve this, researchers also looked into attention mechanisms to merge or replace these models. This has led to trying out Transformer architectures in the computer vision domain which were successful in NLP.

Self-attention-based architectures have taken priority in NLP by avoiding the drawbacks such as vanishing gradients in sequence modeling and transduction tasks. Specially designed for sequence modeling and transduction tasks, Transformers with attention were able to model long-range sequences of data. When training a NLP model, one of the best ways is to pre-train on a large text corpus and then fine-tune on a small set of data which is for the related task. But with deep neural networks, this was a challenging task. As Transformers have high computational efficiency and scalability, it was easier to train on a large set of data \cite{hochreiter1998vanishing}.

With the success of using self-attention to enhance the input-output interaction in NLP, works have been proposed to combine convolutional architectures with self-attention, especially in object detection and semantic segmentation where input-output interaction is highly needed \cite{ramachandran2019stand}. But applying attention to convolutional architectures demands high computation power, even though they are theoretically efficient \cite{Vaswani2017}.

\begin{figure}[htp]
    \centering
    \includegraphics[width=\columnwidth]{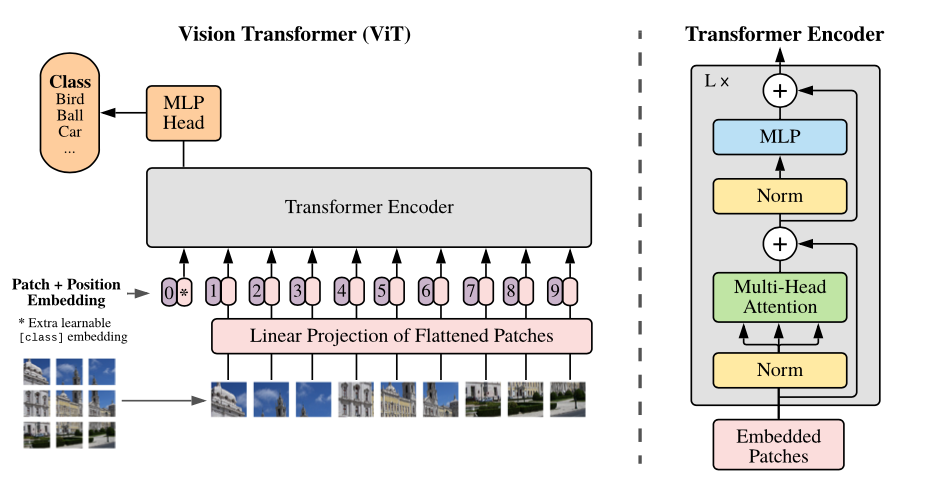}
    \caption{Architecture of the Vision Transformer. The model splits an image into a number of fixed-size patches and linearly embeds them with position embeddings (left). Then the result is fed into a standard transformer encoder (right). Adapted from \cite{Dosovitskiy2020}.}
    \label{fig:vit_figure}
\end{figure}

 Considering images, calculating self-attention is quadratic to the image size as each pixel attends to every other pixel therefore it is a quadratic cost of the pixel count \cite{Dosovitskiy2020}. Thus \cite{Dosovitskiy2020} proposed to divide the image into a sequence of patches and treat them as tokens as it was done in NLP. Instead of pixel-wise attention, patch-wise attention was used in the architecture which helped to reduce the computational complexity compared to applying self-attention to convolutional architecture.
 
 This architecture showed promising results by surpassing all the state-of-the-art convolution-based methods by reaching an accuracy of 88.55\% on ImageNet, 90.72\% on ImageNet-ReaL, and 94.55\% on CIFAR-100 datasets \cite{Dosovitskiy2020}. A major characteristic of the ViT is that it needs more data for model training. Experiments carried out by \cite{Dosovitskiy2020} ensure that with increasing data size, ViT performs well. 

\subsection{Applications of Semantic Segmentation}

In this section, we discuss various application domains of semantic segmentation, including remote sensing, medical imaging, and video processing. For each of these domains, we highlight the unique challenges and opportunities that arise, as well as the current state-of-the-art methods and techniques.

\subsubsection{Semantic Segmentation of Remote Sensing Images}
Remote sensing is the process of getting information and monitoring the characteristics of an area without having any physical contact. The two main types of remote sensing techniques are the use of active sensors such as RADAR, LiDAR and the use of passive sensors such as satellite imagery \cite{zhu2018review}. These high-resolution earth surface images provide a wide range of use cases such as world mapping updates \cite{schmitt2020weakly}, forest degradation analysis \cite{olander2008reference}, monitoring changes to the surface \cite{pacifici2007innovative}, etc.

Remote sensing imagery is widely used in combination with computer vision and Artificial Intelligence (AI) for analyzing and processing the earth's surface over large areas with complex feature distributions. The images collected by satellites or unmanned aerial vehicles (UAV) provide a wide range of information for applications such as urban planning, disaster management, traffic management, climate change, wildlife conservation, crop monitoring, etc. The use of datasets containing these high-resolution images and their respective segmented masks \cite{boguszewski2021landcover} have provided a base for remote sensing image analysis using computer vision and AI. The use of neural networks provides the ability to process large amounts of image data for object detection, semantic segmentation, and change detection tasks. The evolution in the remote sensing domain has further improved satellite sensors and the introduction of drone technology for aerial imagery has been vital to getting finer details on the earth's surface. This has resulted in precise and accurate data for processing using AI techniques \cite{osco2021review}.

Remote sensing images of the earth's surface provide land cover areas that can be categorized into different segmented classes. Each of these classes is assigned a label for each pixel while preserving the spatial resolution of the image. Many datasets containing these remote sensing images and their segmented masks are available  \cite{boguszewski2021landcover,wang2021loveda,demir2018deepglobe} to use for different applications such as change detection, land cover segmentation, and classification. Examples of common land cover classes covered by the pixel-level classification are forests, crops, buildings, water resources, grasslands, roads, etc. Research has been conducted using ViT architecture models by adding layers and attention mechanisms efficiently and improvements in performance to process high-resolution remote sensing images for semantic segmentation such as Efficient Transformer \cite{xu2021efficient} and Wide-Context Transformer \cite{ding2022looking}.

Manual segmentation of these different environmental areas from a complex satellite or aerial images is a difficult task which is time-consuming, error-prone, and requires expertise in the remote sensing domain.

\subsubsection{Semantic Segmentation of Medical Images}

Medical image analysis has developed and incorporated scanning and visualization techniques. Segmentation techniques have been vital as it has the ability to identify and segment medical imagery to assist in further diagnosis and interventions. By identifying each region of interest (ROI) highlighted, various important diagnoses are happening such as brain tumor boundary detection from MRI images, pneumonia affections in X-rays, cancer detection from biopsy sample images, etc. The demand for this type of analysis through image segmentation has emerged in the recent past with much research being done in the scope to develop more precise, efficient models and algorithms. These medical images that are used in image segmentation tasks can be grouped based on modalities such as MRI, CT scan, X-ray, ultrasound, microscopy, dermoscopy, etc. Each of these categories contains datasets that were collected under medical supervision and some are made publicly available.

Since there exist several modalities as mentioned above, the technological systems that are used for medical imagery differ. Medical imagery system development vendors built them as per the doctor's requirements. Therefore, the images generated are bound to the limitations of the technology available and require medical personal intervention to examine them \cite{olabarriaga2001interaction}. Therefore the segmentation of these images in different biological domains requires experts in each field to cope with these systems and spend a vast amount of time examining them. To overcome these difficulties, the capability of automatic feature extraction has been introduced with deep learning based techniques, which have been valuable in the sense of medical imagery. With the advancements in segmentation analysis, better-performing models have been introduced with the use of medical images by many researchers. One such famous architecture is the U-Net \cite{ronneberger2015u} which was initially introduced for medical image analysis. Based on this, several improved versions have been followed up using medical imagery datasets from heart, lesion, and liver segmentation \cite{gu2019net,huang2020unet,zhou2018unet++}. This proves how beneficial the improvement of segmentation has been in the medical environment. In recent years, the emerging new architectures of ViTs have also been applied to the medical domain with TransUNet \cite{chen2021transunet} and Swin-Unet \cite{cao2023swin}. They are hybrid Transformer architectures with the advantages of the U-Net. They performed with better accuracy in cardiac and multi-organ segmentation applications.

Some limitations of medical images are the relatively less number of images available compared to natural image datasets (landscapes, people, animals, and automobiles) with millions of images. In the medical domain, there are several image modalities. For annotating medical images, expertise in each medical field is a must. Among them, MRI and microscopy images are quite difficult to annotate \cite{icsin2016review}. Typically, these datasets contain fewer images compared to ultrasound, X-ray, and lesion datasets which are obtained with the existing scanning systems and are easier to annotate with less complex structures and fine boundaries. But still, limitations exist due to restrictions on privacy and other medical policies to obtain these images in large quantities. To overcome these limitations with some datasets, several image segmentation challenge competitions are taking place every year which provide publicly available well-annotated medical image datasets. Most of the improvements made through research in semantic segmentation models have been based on these challenge datasets and most are taken as benchmark datasets for segmentation \cite{codella2018skin,bilic2019liver,menze2014multimodal}.

\subsubsection{Video Semantic Segmentation}
Human-Machine interaction \cite{gorecky2014human}, augmented reality \cite{azuma1997survey}, autonomous vehicles \cite{janai2020computer}, image search engines \cite{gevers2004image} are some applications in complete scene understanding and for these type of applications, semantic segmentation contributes more on complete scene understanding on videos. Usually, the idea is to apply semantic segmentation on frames of a high-resolution video where the video is considered as a set of uncorrelated fixed images \cite{jain2019accel}. The common challenge with this type of semantic segmentation is the computational complexity of scaling the spatial dimension of the video using the temporal frame rate. Removal of temporal features and only focusing on spatial frame-by-frame features doesn't make sense in video segmentation. Since there is a combined flow among frames of a video, considering the temporal context of a video is an essential factor in video semantic segmentation, even though it is computationally expensive.

Research has been conducted to reduce this high computation cost on videos. Feature reuse and feature warping \cite{ding2020every} have been proposed as a solution. Cityscapes \cite{cordts2016cityscapes} and CamVid \cite{brostow2009semantic}, are some largest video segmentation datasets available for frame-by-frame approach of video segmentation \cite{richter2016playing}. Recent papers have proposed segmentation methods such as selective re-execution of feature extraction layers \cite{shelhamer2016clockwork}, optical flow-based feature warping \cite{zhu2017deep}, and LSTM-based, fixed-budget keyframe selection policies \cite{mahasseni2017budget}. The main key problem in these approaches is that they have less attention to the temporal context of a video. Researchers have shown that to satisfy both spatial and temporal contexts, using an optical flow of video as temporal information to speed up uncertainty estimation makes good sense \cite{garcia2018survey}. VisTR \cite{wang2021end}, TeViT \cite{yang2022temporally} and SeqFormer \cite{wu2022seqformer} are some of the Transformer models that are used for video segmentation tasks.

\subsection{Practical approaches to overcome the data limitation}

Deep neural networks have performed well with supervised learning in computer vision and NLP. But when it comes to the real world, supervised learning faces a bottleneck in training a neural network as it needs lots of labeled data. Collecting labeled data or manual labeling is difficult in every aspect. Training a network from scratch is a somewhat costly task; as a remedy for this, transfer learning comes into play. But when considering specified downstream tasks such as satellite imagery semantic segmentation, using pre-trained datasets is difficult as most of the architectures have been trained on benchmark datasets where the data domain is different. Therefore, getting good accuracy has been tricky. 

\begin{figure}[htp]
    \centering
    \includegraphics[width=\columnwidth]{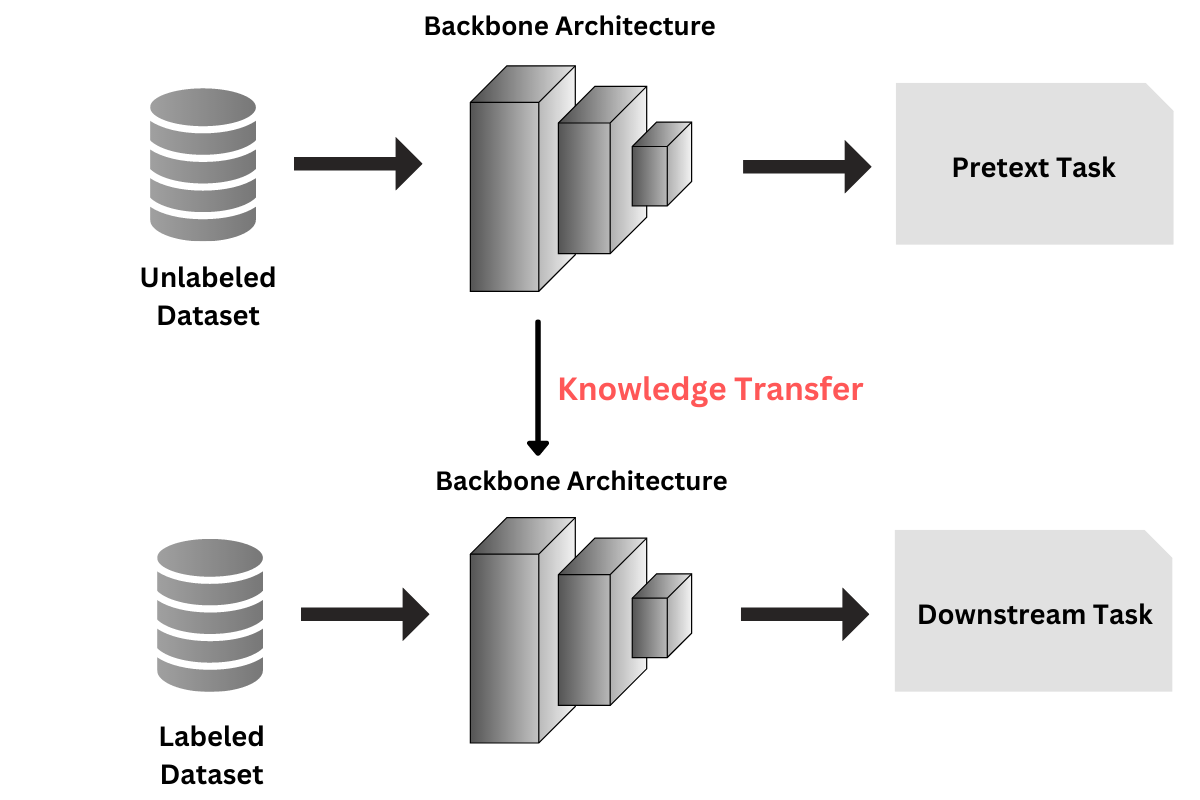}
    \caption{The general pipeline of self-supervised learning. The trained weights from solving a pretext task are applied to solve some downstream tasks.}
    \label{fig:sslpre}
\end{figure}

Specially when considering Transformer architectures, self-supervised learning plays a great role as a remedy for data-hungry problems in deep learning. In human vision, humans are fed with different things in the environment and then are able to distinguish those things from other objects in the environment. There are no labeling mechanisms for these scenarios. Therefore, this is the technique used in SSL which actually trains a neural network using an unlabeled dataset where the labels are automatically provided through the dataset itself. As the first step, the network is set to solve a pretext task as described in Figure \ref{fig:sslpre}. A pretext task is a pre-designed task from which the network can learn features and then using those trained weights for different features, the network can be applied to solve some downstream tasks. A downstream task is a specified task. Common downstream tasks in computer vision are semantic segmentation, object detection, etc.  

Rotating an image by a given angle and predicting the rotation, solving jigsaw puzzles, filling a cut patch on an image, predicting the relative position of a patch of an image, and separating images belonging to different clusters can be considered as some of the pretext tasks in SSL \cite{gustavsson2021object}. By using these methods, the network can learn different features in the dataset under the given scope. No labels are used here and automatic labeling is achieved via the image itself.

SSL has three general categories based on how the training happens.
\begin{itemize}
\item Generative: Train the encoder to encode the given input and using the decoder get the input back
\item Contrastive: Train the encoder to encode the given input and find the similarities
\item Generative-Contrastive (Adversarial): Train encoder to encode the given input and create fake outputs and compare the features of the input and output \cite{liu2021self}
\end{itemize}

Semantic segmentation is one of the major downstream tasks that can be performed using SSL. Pixel-wise labeling is essential in semantic segmentation. If there are no properly annotated datasets, SSL is the best way to train semantic segmentation architectures. 

\subsection{Loss functions in semantic segmentation}

For segmentation, classification, and object detection models accuracy improvement not only depends on the model architectures but also on the loss functions used. The loss function calculates the overall error while training batches and adjust the weights through back propagation.  
Numerous loss functions have been created to cope with various domains, and some of them are derived from existing loss functions. Additionally, these loss functions take into account the imbalances in the dataset too.

In the case of semantic segmentation, the default choice and most commonly used is the cross-entropy loss which is applied pixel-wise. The loss function independently evaluates the class predictions for each pixel and averages over all the pixels.
\begin{equation}
    CE_{loss}(p, q) = -\sum_{i=1}^{n} p_i \log(q_i)
    \label{eq:cross-entropy}
\end{equation}

The equation \ref{eq:cross-entropy} above computes the average loss for each pixel in an image. Here in the equation $p_i$ is the true probability of the $i^{th}$ class and $q_i$ is the predicted probability of the same class. This supports the model to generate probability maps that closely resemble the actual segmentation masks while penalizing inaccurate predictions more heavily. By minimizing the cross-entropy loss function during training, the model becomes better at precise image segmentation.

Even though the above method is widely used it can be biased with dataset imbalance as the majority class will be dominant. To overcome this when the dataset is skewed, a weighted cross entropy loss is introduced in \cite{ronneberger2015u}.
\begin{equation}
    WCE_{loss}(p, q) = -\sum_{i=1}^{n} p_i w_i \log(q_i)
    \label{eq:cross-entropy-weighted}
\end{equation}
Here as in equation \ref{eq:cross-entropy-weighted}, a weight factor as $w_i$ for the $i^{th}$ class is inserted to the typical equation \ref{eq:cross-entropy}. 
But the issue was not significantly solved as the cross entropy calculates the average per-pixel loss without considering the adjacent pixels which can be boundaries.

As a further improvement for the cross-entropy loss, the focal loss technique \cite{lin2017focal} was introduced. This is implemented by altering the structure of cross-entropy loss. When focal loss is applied to samples with accurate classifications, the scaling factor value is down-weighted. This ensures the more harder samples are emphasized, therefore high class imbalance won’t bias toward the overall calculations. 

\begin{equation}
    F_{loss}(p_t) = -\alpha_t (1 - p_t)^\gamma \log(p_t)
    \label{eq:focal-loss}
\end{equation}
In the equation \ref{eq:focal-loss}, $p_t$ is the predicted probability of the true class, $\alpha_t$ is a scaling factor that gives higher weight to the positive class, and $\gamma$ is a focusing parameter that controls how much the loss is focused on hard examples. 

The cross-entropy loss is scaled in this loss function, with the scaling factors decreasing to zero as the confidence in the well-classified classes rises. Therefore more attention is given to the pixel classes which are difficult to predict.

Another set of loss calculation techniques is the overlapping between prediction and actual segmentations. The models are trained to minimize the loss such that the model outputs segmentations with higher overlaps. 

Dice loss is one such widely used popular measure in computer vision tasks to calculate the similarity between two images. It is based on the dice coefficient which was later developed as the dice loss function in the segmentation domain. This loss was first used in the computer vision domain by \cite{milletari2016v} in medical image segmentation tasks.

\begin{equation}
D_{loss}(g, p) = 1 - \frac{2 \sum_{i=1}^{n} g_i p_i}{\sum_{i=1}^{n} g_i + 
\sum_{i=1}^{n} p_i + \epsilon}
 \label{eq:dice-loss}
\end{equation}

Here, in equation \ref{eq:dice-loss} $g$ and $p$ describes the ground truth and prediction segmentations. The sum is calculated over the $n$ number of pixels with $\epsilon$ small constant added to avoid division by zero. The dice coefficient measures the overlap between the samples (ground truth and prediction) and provides a score ranging from 0 to 1, 1 means perfect overlap. Since this method considered pixels in both global and local contexts, the accuracy is higher than cross-entropy loss calculations.

Another similar method used to evaluate the metric of models is the IoU (Intersection over Union) loss also known as the Jaccard index. It is quite similar to the dice metric and measures the overlapping of the positive instances between the considered samples. This method as shown in equation \ref{eq:iou-loss} differs from the 
dice loss with correctly classified segments relative to total pixels in either the ground truth or predicted segments.

\begin{equation}
    IoU_{loss} (g,p) = 1 - \frac{\sum_{i=1}^{n} g_i p_i}{\sum_{i=1}^{n} g_i + \sum_{i=1}^{n} p_i - \sum_{i=1}^{n} g_i p_i + \epsilon}
    \label{eq:iou-loss}
\end{equation}

For multi-class segmentation, the mean IoU is considered by taking the average of each individual class IoU. This is widely used for performance comparison and evaluation of dense prediction models \cite{jadon2020survey}. 

\section{Datasets} \label{Datasets}

In this section, the common datasets used for the training and testing of semantic segmentation models are considered. Factors affecting the creation of  real datasets are lighting conditions, weather, and season. Based on these factors, datasets can be classified into different groups. When data is collected under normal daytime environmental conditions, those data are categorized under no cross-domain datasets. If data is collected under some deviated environmental conditions including rainy, cloudy, nighttime, snowy, etc then such data are categorized under cross-domain datasets. Another category is synthetic data, where the data is artificially created and collected for training purposes. These synthetic datasets are mostly created as a cost-effective supplement for training purposes. Following are some of the benchmark datasets specially made for semantic segmentation tasks, with a summary presented in Table \ref{table:dataset-summary}. 

\textbf{PASCAL-Context} \cite{mottaghi2014role} This dataset was created by manually labeling every pixel of PASCAL-VOC 2010 \cite{everingham2012pascal} dataset with semantic categories. The domain of this dataset is not limited and its data contains different objects. The semantic categories of this dataset can be divided into three main classes. (i) objects, (ii) stuff, and (iii) hybrids. Objects have defined categories such as cups, keyboards, etc. Stuff has classes without a specific shape and has regions such as sky, water, etc. Hybrid contains intermediate objects such as roads where roads have a clear boundary but shape cannot be predicted correctly. 

\textbf{ADE20K} \cite{zhou2017scene} Annotations of this dataset are done on scenes, objects, parts of objects. Many of the objects in the dataset are annotated with their parts. Annotations in this dataset are made continuously. Therefore, this is a growing dataset. 

\textbf{KITTI} \cite{kuutti2020survey} This dataset contains both 2D and 3D images which have been collected from urban and rural expressway incidents and traffic scenarios. It is useful for robotics and autonomous driving. This dataset has different variants namely KITTI-2012, KITTI-2015 and they have some differences in the ground truth. 

\textbf{Cityscapes} \cite{cordts2016cityscapes} This contains large-scale pixel-level and instance-level semantic segmentation annotations recorded from a set of stereo video sequences. Compared to other datasets, quality, data size, and annotations in this dataset have a good rank and data have been collected from 50 different cities in Germany and neighboring countries. 

\textbf{IDD} \cite{varma2019idd} This is specially designed for road scene understanding and data have been collected from 182 Indian road scenes. As these are taken from Indian roads, there are some variations in the weather and lighting conditions because of dust and air quality on roads. One key feature of this dataset is, this contains some special classes such as auto-rickshaws and animals on the roads. 

\textbf{Virtual KITTI} \cite{gaidon2016virtual} Except for different weather and imaging conditions, most of the virtual vision datasets such as Virtual KITTI are similar to the real vision datasets. Therefore virtual datasets are useful for pre-training purposes. This dataset is created from 5 different urban scene videos from the real-world KITTI dataset. Data have been automatically labeled and can be used for object detection, semantic segmentation, instance segmentation, etc.

\textbf{IDDA} \cite{alberti2020idda} This contains 1 million frames generated from simulator CARLA oriented on different 7 city models. This dataset can be used to do semantic segmentation for more than 100 different visual domains and is specially designed for autonomous driving models.

\begin{table*}[h]
  \centering
  \resizebox{\columnwidth}{!}{%
    \begin{tabular}{ c c c c c c c c}
     \hline
     Dataset & Classes & Size & Train & Validation & Test & Resolution (pixels) & Category\\
     \hline 
     PASCAL-Context & 540 & 19740 & 4998 & 5105 & 9637 & $387 \times 470$ & No cross-domain \\
     ADE20K         & 150 & 25210 & 20210 & 2000 & 3000 & - & No cross-domain \\
     KITTI          & 5 & 252 & 140 & - & 112 & $1392 \times 512$ & No cross-domain\\
     Cityscapes     & 30 & 5K fine, 20K coarse &  2975 & 500 & 1525 & $1024 \times 2048$ & Cross-domain\\
     IDD            & 34 & 10004 & 7003 & 1000 & 2001 & $1678 \times 968$ & Cross-domain\\
     Virtual KITTI  & 14 & 21260 & - & - & - & $1242 \times 375$ & Synthetic\\
     IDDA           & 24 & 1M & - & - & - & $1920 \times 1080$ & Synthetic\\
     \hline
    \end{tabular}%
    }
    \caption{\textbf{Summary of the datasets} \\ Note: Both cross-domain and no-cross domain falls into the non-synthetic category}
    \label{table:dataset-summary}
\end{table*}

\section{Meta - analysis} \label{Meta - analysis}

In this section, we discuss some of the ViT models specialized for the task of semantic segmentation. The models are selected upon considering the datasets that they benchmarked (ADE20K, Cityscapes, PASCAL-Context). The intuition behind that is to compare all the models on a common basis. The benchmark results are summarized in Table \ref{table:benchmark-results}. 
 
\subsection{\textbf{SE}gmentation \textbf{TR}ansformer (SETR)}

SETR \cite{zheng2021rethinking} proposes semantic segmentation as a sequence-to-sequence prediction task. They adopt a pure Transformer as the encoder part of their segmentation model without utilizing any convolution layers. In this model, they replace the prevalent stacked convolution layer based encoder with a pure Transformer which gradually reduces the spatial resolution. 

\begin{figure}[htp]
    \centering
    \includegraphics[width=\columnwidth]{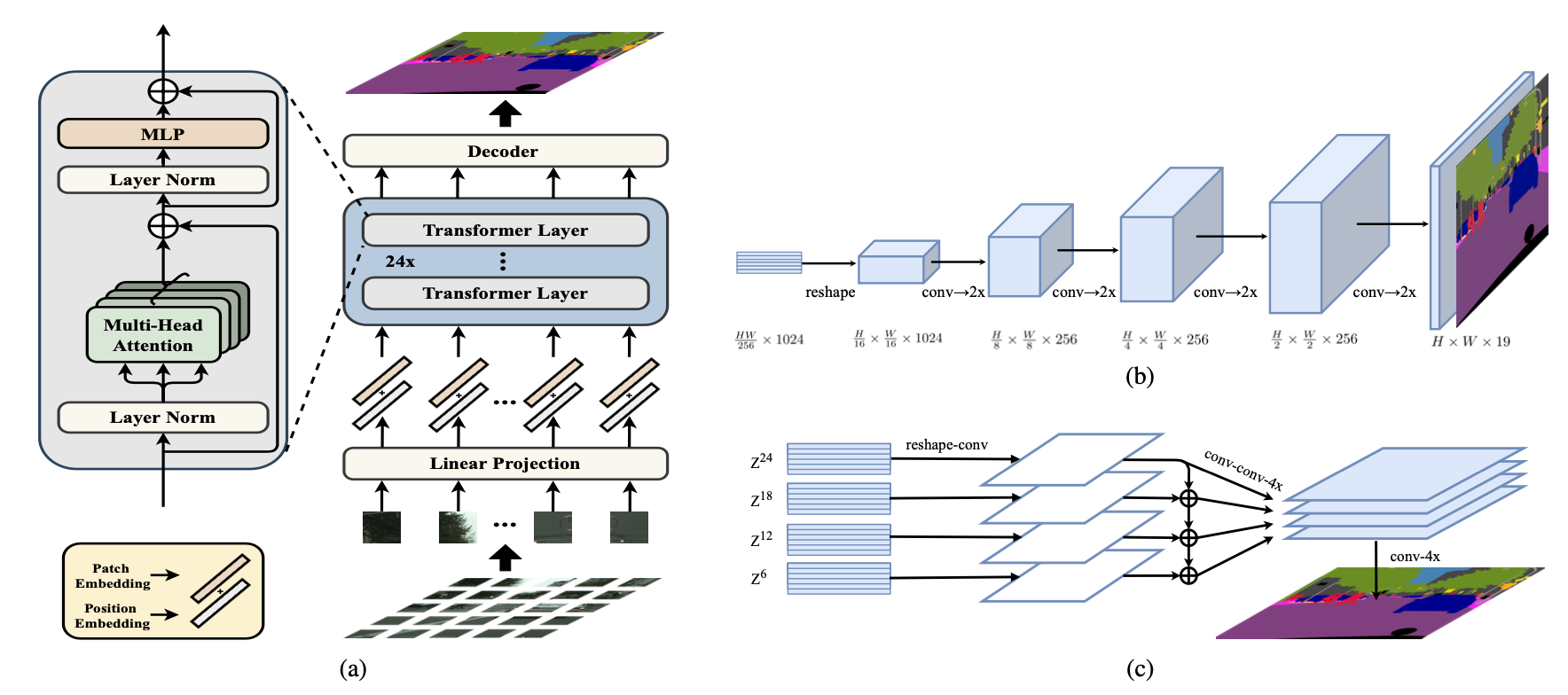}
    \caption{SETR architecture and its variants adapted from \cite{zheng2021rethinking}. (a) SETR consists of a standard Transformer. (b) \textit{SETR-PUP} with a progressive up-sampling design. (c) \textit{SETR-MLA} with a multi-level feature aggregation.}
    \label{fig:setr}
\end{figure}

The SETR encoder (Figure \ref{fig:setr}a) which is a standard Transformer treats an image as a sequence of patches followed by a linear projection. Then it embeds these projections with patch embedding + position embedding to feed them into a set of Transformer layers. SETR has no down-sampling  in spatial resolution at each layer of the encoder transformer while it only provides global context modeling.  They classify SETR into a few variants depending on the decoder part of the model; \textit{SETR-PUP} (Figure \ref{fig:setr}b) which has a progressive up-sampling design and the \textit{SETR-MLA} (Figure \ref{fig:setr})which has a multi-level feature aggregation.  

SETR achieved state-of-the-art semantic segmentation results on ADE20K, Pascal Context by the time of submission \cite{zheng2021rethinking}. It has also been tested on the Cityscapes dataset and has shown impressive results. 

\subsection{Swin Transformer}

To address the issue of not having a general purpose Transformer backbone for computer vision tasks, \cite{liu2021swin} proposed Swin Transformer (Hierarchical Vision Transformer using \textbf{S}hifted \textbf{Win}dows) which can be served as a general purpose backbone for computer vision tasks such as image classification and dense prediction.  

\begin{figure}[htp]
    \centering
    \includegraphics[width=\columnwidth]{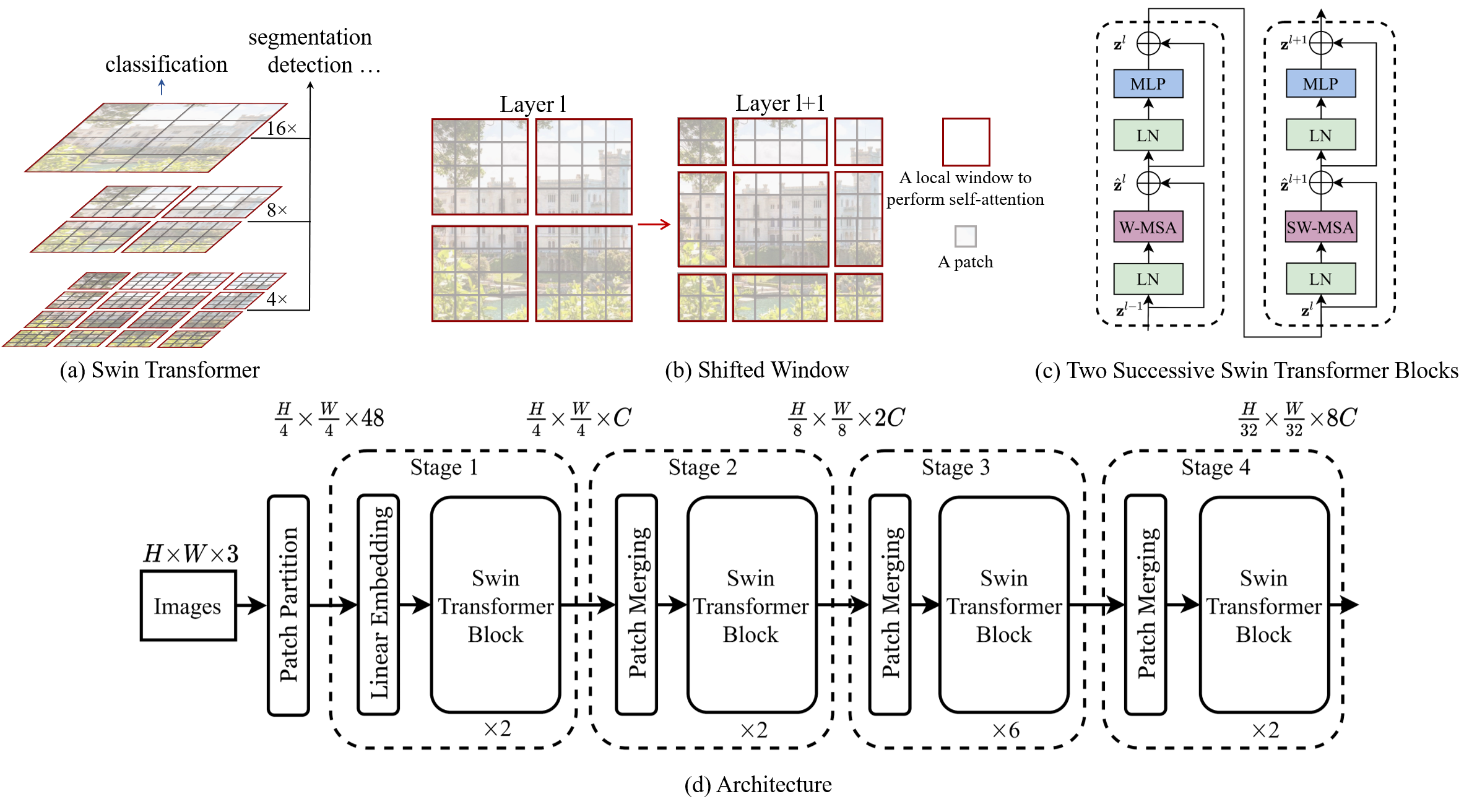}
    \caption{An overview of the Swin Transformer adapted from \cite{liu2021swin}. (a) Hierarchical feature maps for reducing computational complexity. (b) Shifted
window approach which was used when calculating self-attention. (c) Two successive Swin Transformer Blocks which presented at each stage. (d) Core architecture of the Swin.}
    \label{fig:swin}
\end{figure}

Swin Transformer was able to bring down the quadratic computational complexity of calculating self-attention in Transformers to linear complexity by constructing hierarchical feature maps (Figure \ref{fig:swin}a). Also, the shifted window approach illustrated in Figure \ref{fig:swin}b has much lower latency than the earlier sliding window based approaches which were used to calculate the self-attention. Swin Transformer showed great success over the previous state-of-the-art in image classification (87.3\% top-1 accuracy on ImageNet-1K), semantic segmentation (53.5\%  mIoU  on  ADE20Kval) and object detection (58.7 box AP and 51.1 mask AP on COCO test-dev) \cite{liu2021swin}. 

According to the architecture of a Swin Transformer, in the beginning, it splits the given image into a sequence of non-overlapping patches (tokens) by using the patch partitioning module (Figure \ref{fig:swin}d). Then a linear embedding is applied to this sequence of patches to project them into an arbitrary dimension. It is followed by several Swin Transformer blocks to apply self-attention. The main responsibility of the patch merging module is to reduce the number of tokens in  deeper layers.  It is noteworthy that the feature map resolutions in the hierarchical stages are similar to those in typical convolution architectures such as ResNet \cite{he2016deep}. Therefore Swin Transformer can efficiently replace ResNet backbone networks in computer vision tasks. 

\subsection{Segmenter}

\begin{figure}[htp]
    \centering
    \includegraphics[width=\columnwidth]{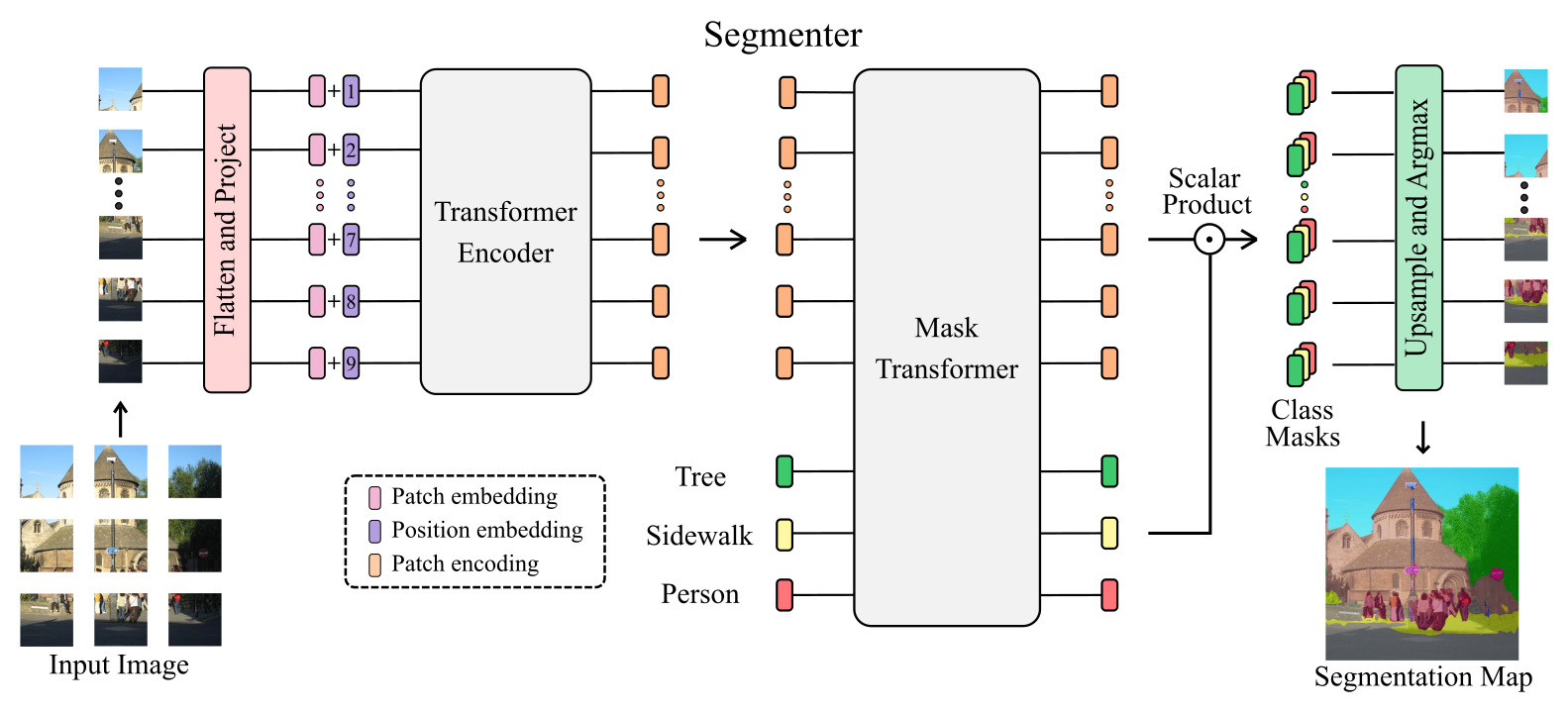}
    \caption{Segmenter architecture adapted from \cite{strudel2021segmenter}. It basically has a ViT backbone with a mask transformer as the decoder.}
    \label{fig:segmenter}
\end{figure}

Segmenter \cite{strudel2021segmenter} is a purely transformer-based approach for semantic segmentation which consist of a ViT backbone pre-trained on ImageNet and introduces a mask transformer as the decoder (Figure \ref{fig:segmenter}). Even though the model was built for segmentation tasks, they take advantage of the models made for image classification to pre-train and then fine-tune them on moderate-sized segmentation datasets. 

CNN-based models are generally inefficient when processing global image context and ultimately result in a sub-optimal segmentation. The reason for the sub-optimal segmentation of the convolution-based approaches is that convolution is a local operation which poorly accesses the global information of the image. But the global information is crucial where the global image context usually influences the local patch labeling. But modeling of global interaction has a quadratic complexity to the image size because it needs to model the interaction between each and every raw pixel of the image. The architecture of the Segmenter especially captures the global context of images, unlike the traditional CNN-based approaches. 

Other than the semantic segmentation tasks, this Segmenter model also can be applied to panoptic segmentation (semantic segmentation + instance segmentation) tasks by altering the model architecture. The class embeddings of the model need to be replaced by object embeddings in such a case.

\subsection{SegFormer}

\begin{figure}[htp]
    \centering
    \includegraphics[width=\columnwidth]{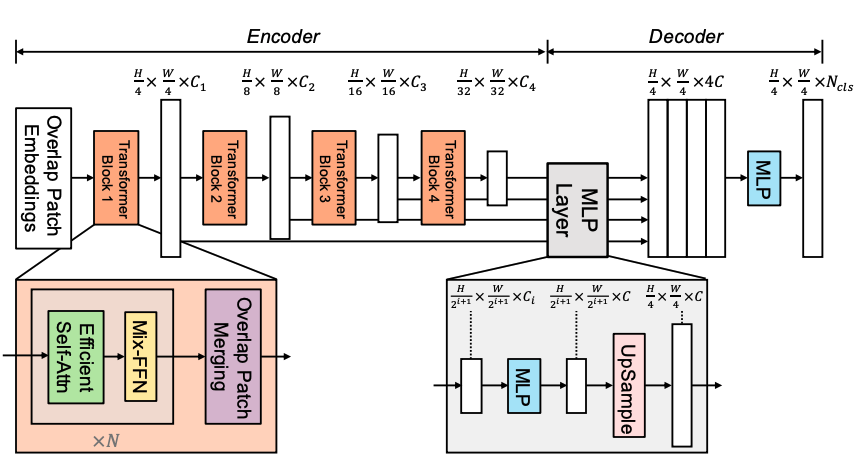}
    \caption{SegFormer architecture adapted from \cite{xie2021segformer}. It has a hierarchical Transformer
encoder for feature extraction and a lightweight MLP decoder for predicting the final mask.}
    \label{fig:segformer}
\end{figure}

SegFormer \cite{xie2021segformer} is an  architecture for semantic segmentation which consist of a hierarchical Transformer encoder with a lightweight multilayer perceptron (MLP) decoder (Figure \ref{fig:segformer}). The MLP decoder is used for predicting the final mask. To obtain a precise segmentation, it uses a patch size of $4 \times 4$ in contrast to ViT which uses a patch size of $16 \times 16$. It has an overlapped patch merging process to maintain the local continuity around the patches. 

Generally, ViT has a fixed resolution for positional encoding \cite{chu2021conditional}. This leads to a drop in accuracy since it needs to interpolate the positional encoding of testing images when they have a different resolution than training images. Thus, SegFormer introduces a Positional-Encoding-Free design as a key feature. 

Moreover, the authors claim their architecture is more robust against common corruptions and perturbations than current methods which make SegFormer appropriate for safety-critical applications. SegFormer achieved competitive results on ADE20K, Cityscapes, and COCO-Stuff datasets as shown in Table \ref{table:benchmark-results}. SegFormer comes in several variants from SegFormer-B0 to SegFormer-B5, where the largest model is SegFormer-B5. This largest model surpasses the SETR \cite{zheng2021rethinking} on the ADE20K dataset achieving the highest mIoU while being $4 \times $ faster than SETR. All of these SegFormer models have trade-offs between model size, accuracy, and runtime. 

\subsection{Pyramid Vision Transformer (PVT)}

ViT couldn't be directly applicable to dense prediction tasks because its output feature map is single scaled and it generally has a low resolution which comes at a higher computational cost. PVT \cite{wang2021pyramid} overcomes the aforementioned concerns by introducing a progressive shrinking pyramid backbone network to reduce the computational costs and simultaneously output more fine-grained segmentation. PVT comes in two variants. PVT v1 \cite{wang2021pyramid} is the first work by the authors and PVT v2 \cite{wang2022pvt} comes with some additional improvements to the previous version. 

\subsubsection{PVT v1}
\begin{figure}[htp]
    \centering
    \includegraphics[width=\columnwidth]{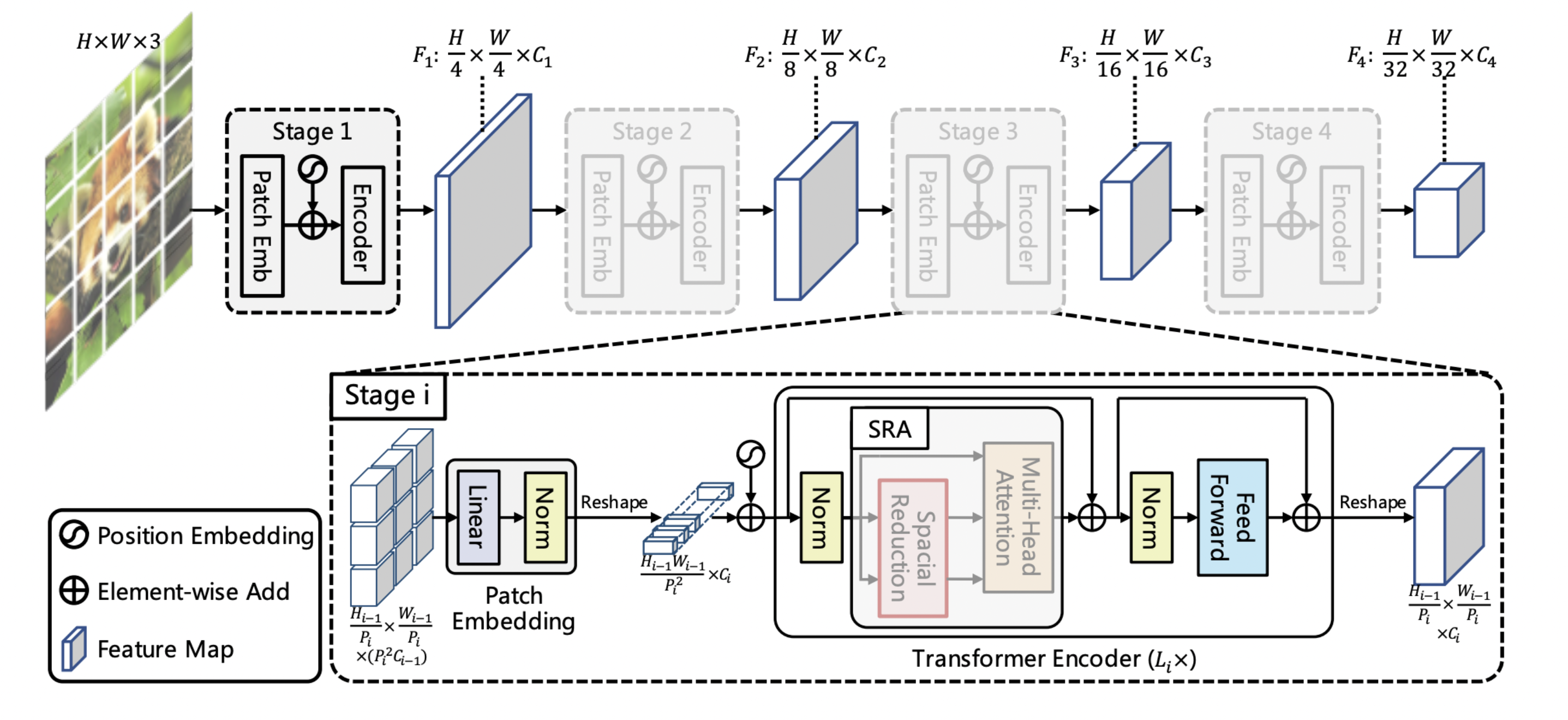}
    \caption{PVT v1 architecture adapted from \cite{wang2021pyramid}. The pyramid structure of the stages progressively shrinks the output resolution from high to low.}
    \label{fig:pvt_v1}
\end{figure}

This initial version has some noteworthy changes compared to the ViT. It takes $4 \times 4$ input patches in contrast to the $16 \times 16$ patches in ViT. This improves the model's ability to learn high-resolution representations. It also reduces the computational demand of traditional ViT by using a progressive shrinking pyramid. This pyramid structure progressively shrinks the output resolution from high to low in the stages which are responsible for generating the scaled feature maps (Figure \ref{fig:pvt_v1}). Another major difference is that it replaces the multi-head attention layer (MHA) in ViT with a novel spatial reduction attention (SRA) layer which reduces the spatial scales before the attention operation. This further reduces the computational and memory demand because SRA has a low computational complexity than MHA. 

\subsubsection{PVT v2}
\begin{figure}[htp]
    \centering
    \includegraphics[width=0.8\columnwidth]{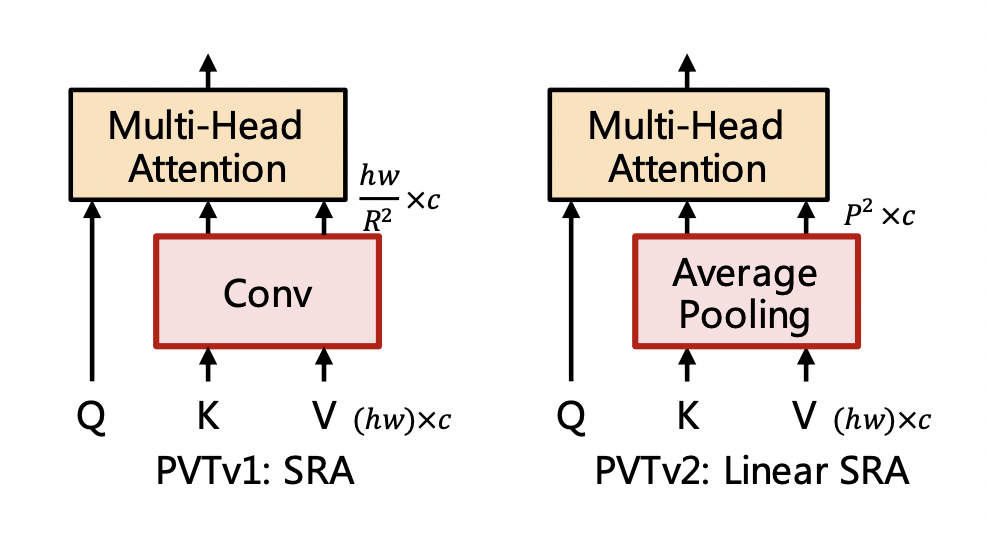}
    \caption{Comparison of spatial reduction attention (SRA) layers in PVT versions \cite{wang2022pvt}}
    \label{fig:pvt_v2_1}
\end{figure}

\begin{figure}[htp]
    \centering
    \includegraphics[width=0.6\columnwidth]{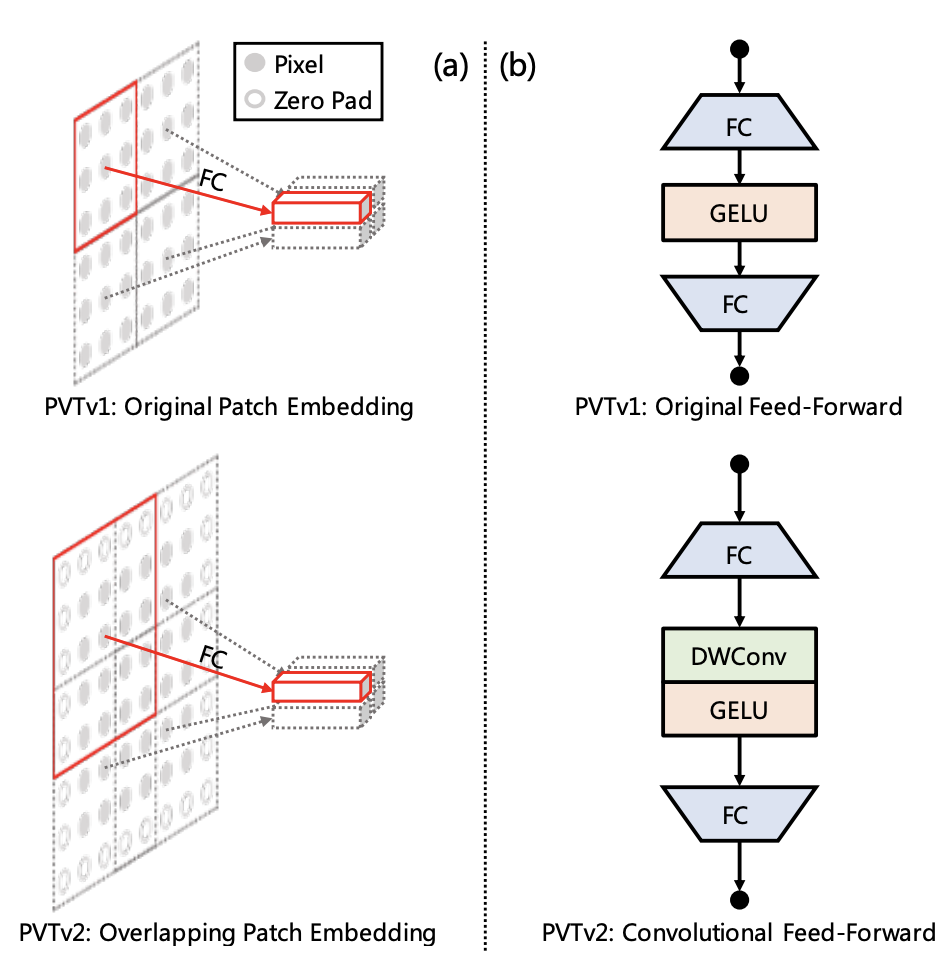}
    \caption{Improved patch embedding and feed-forward networks in PVT v2 \cite{wang2022pvt}}
    \label{fig:pvt_v2_2}
\end{figure}

The former version has a few drawbacks. The computational demand of the PVT v1 is relatively large when processing high-resolution images. It loses the local continuity of the images when processing the image as a sequence of non-overlapping patches. It cannot process variable-sized inputs because of the fixed-size position encoding.  
This new version has three major improvements which circumvent the previous design issues. First one is linear spatial reduction attention (LSRA) which reduces the spatial dimension of the image to a fixed size using average pooling (Figure \ref{fig:pvt_v2_1}). Unlike SRA in the PVT v1, LSRA benefits from linear complexity. Second one is the overlapping patch embedding (Figure \ref{fig:pvt_v2_2}a). This is done by zero-padding the border of the image and taking more enlarged patch windows which overlap with the adjacent windows. It helps to capture more local continuity of the images. The third one is the convolutional feed-forward network (Figure \ref{fig:pvt_v2_2}b) which helps to process different sizes of input resolutions. With these major improvements, PVT v2 was able to bring down the complexity of PVT v1 to linear complexity. 

We can clearly see how the improvements of the PVT v2 contribute to higher gains in the benchmark comparison in Table \ref{table:benchmark-results}. 

\subsection{Twins}
Twins \cite{chu2021twins} propose two modern Transformer designs for computer vision named Twins-PCPVT and Twins-SVT by revisiting the work on the PVT v1 \cite{wang2021pyramid} and Swin Transformer \cite{liu2021swin}.

Twins-SVT uses a spatially separable self-attention (SSSA) mechanism based on the depth-wise separable convolutions in neural networks. This SSSA has two underlying attention mechanisms which are capable of capturing local information as well as global information. Locally grouped self-attention (LSA) and global sub-sampled attention (GSA) are the above-mentioned attention mechanisms respectively. Those techniques greatly reduce the heavy computational demand in high-resolution image inputs while keeping a fine-grained segmentation. 

\begin{figure}[htp]
    \centering
    \includegraphics[width=\columnwidth]{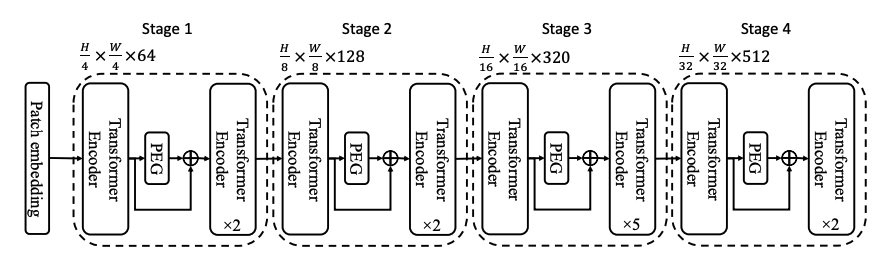}
    \caption{Twins-PCPVT architecture adapted from \cite{chu2021twins}. It uses conditional position encoding with a positional encoding generator (PEG) to overcome some of the drawbacks of fixed-positional encoding.}
    \label{fig:twins}
\end{figure}

As we discussed in the Pyramid Vision Transformer section, PVT v1 can only process fixed-size image inputs due to its absolute positional encoding. This hinders the performance of PVT. To alleviate this challenge Twins-PCPVT uses a conditional position encoding (CPE) first introduced in Conditional Position encoding Vision Transformer (CPVT) \cite{chu2021conditional}. This is illustrated as the positional encoding generator (PEG) in Figure \ref{fig:twins}. It is capable of alleviating some of the issues encountered in fixed-position encoding.  

Twins architectures have shown outstanding performance on computer vision tasks including image classification and semantic segmentation. The semantic segmentation results achieved by the two Twins architectures are highly competitive compared to the Swin Transformer \cite{liu2021swin} and PVT \cite{wang2021pyramid}. 

\subsection{Dense Prediction Transformer (DPT)}
DPT \cite{ranftl2021vision} architecture is introduced with a transformer backbone inside the encoder-decoder design for fine-grained output segmentation predictions compared to the fully convolutional networks. The transformer encoder based on ViT \cite{Dosovitskiy2020} is capable of maintaining spatial resolution over all the stages of the Transformer architecture which is important for dense predictions.

\begin{figure}[htp]
    \centering
    \includegraphics[width=\columnwidth]{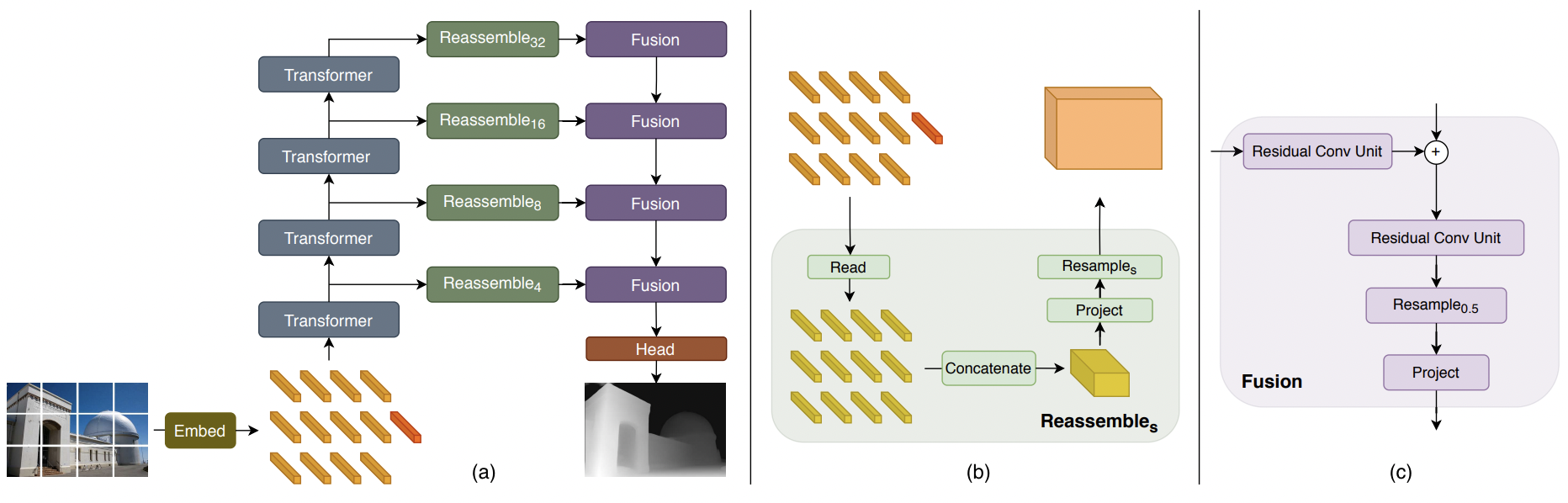}
    \caption{DPT architecture adapted from \cite{ranftl2021vision}. (a) Non-overlapping image patches are fed into the Transformer block. (b) Reassemble operation for assembling tokens into feature maps. (c) Fusion blocks for combining feature maps. }
    \label{fig:dpt}
\end{figure}

In the paper, the authors have introduced several models based on the used image embedding technique. The DPT-Base and DPT-Large models use patch-based embedding where the input image is separated into non-overlapping image patches. Then these are fed into the Transformer block with a learnable position embedding to locate the spatial position of each individual token (Figure \ref{fig:dpt}a). DPT-Base has 12 transformer layers compared to the DPT-Large which has 24 layers with wide feature sizes. The other model is the DPT-Hybrid, which uses the convolutional backbone ResNet-50 as a feature extractor and uses the pixel-based feature maps as token inputs to the 12-layer transformer block.  The Transformer blocks reassemble the tokens with multi-head self-attention (MSA) \cite{Vaswani2017} sequential blocks for global interaction between tokens. The tokens are reassembled into image-like feature representations in various resolutions (Figure \ref{fig:dpt}b). Finally, these representations are combined using residual convolutional units in the decoder and fused together for the final dense prediction (Figure \ref{fig:dpt}c).

The experimental results of the dense prediction transformer have provided improved accuracy results over several benchmark dataset comparisons. The results show that for a large training dataset, the model has the best performance. The comparisons were done for depth estimations and semantic segmentation. ADE20K dataset is used for segmentation and the DPT-Hybrid model has outperformed all the fully-convolutional models \cite{ranftl2021vision}. The DPT has the ability to identify precise boundaries of objects with less distortion. The DPT model was also compared with the PASCAL-Context dataset after fine-tuning.

\subsection{High-Resolution Transformer (HRFormer)}
HRFormer \cite{yuan2021hrformer} is an architecture model that is built using a depth-wise convolutional design with a Feed Forward Network (FFN) and a local window self-attention mechanism with a multi-resolution parallel transformer module. This model is developed for dense prediction tasks focusing on pose estimation and semantic segmentation. The model outperforms the conventional ViT model which produces low-resolution outputs. The HRFormer is designed to maintain the high-resolution using multi-resolution streams and is more efficient in computational complexity and memory usage.



\begin{figure}[htp]
    \centering
    \includegraphics[width=\columnwidth]{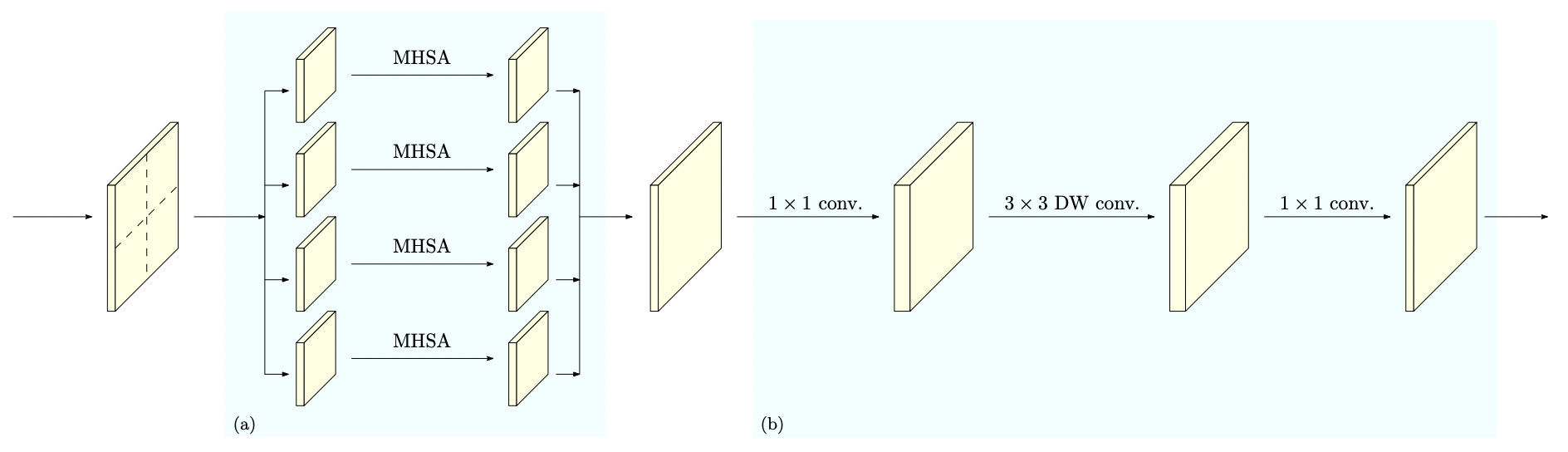}
    \caption{HRFormer architecture adapted from \cite{yuan2021hrformer}. (a) Self-attention blocks. (b) FFN with depth-wise convolutions.}
    \label{fig:hrformer}
\end{figure}

HRFormer has been incorporated by using the HRNet \cite{wang2020deep}, which is a convolutional network consisting of a multi-scale parallel design. This architecture helps to capture feature maps in variant resolutions while maintaining high resolution. At each of these resolution blocks, partitioning is done by creating non-overlapping windows, and self-attention is performed on each image window separately. This improved the efficiency significantly compared to overlapping local window mechanisms introduced earlier in different studies \cite{hu2019local}. The self-attention blocks (Figure \ref{fig:hrformer}a) are followed by an FFN with depth-wise convolutions (Figure \ref{fig:hrformer}b) to increase the receptive field size by information exchange between local windows, which is vital in dense prediction. By incorporating a multi-resolution parallel transformer architecture with convolutional multi-scale fusions for the overall HRFormer architecture, the information between different resolutions is exchanged repeatedly. This process creates a high-resolution output with both local and global context information.

\subsection{Masked-attention Mask Transformer (Mask2Former)}

Mask2Former \cite{cheng2022masked} is a new transformer architecture that can be leveraged to do segmentation tasks including panoptic, instance, and semantic segmentation. It is a successful attempt to introduce a universal architecture for the segmentation tasks which outperforms the current specialized SOTA architectures for each of the segmentation tasks by the time of submission. Its key components consist of a transformer decoder with masked attention. Generally, a standard Transformer attends to the full feature map. In contrast, the masked attention operator in Mask2Former restricts the cross-attention to the foreground region of the predicted mask and then extracts the localized features. This makes the attention mechanism more efficient in this model. 

\begin{figure}[htp]
    \centering
    \includegraphics[width=0.7\columnwidth]{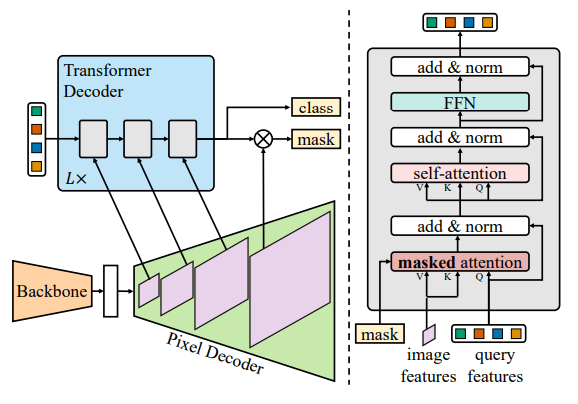}
    \caption{Mask2Former architecture adapted from \cite{cheng2022masked}. The model consists of a backbone feature extractor, a pixel decoder, and a Transformer decoder.}
    \label{fig:mask2former}
\end{figure}

The architecture of Mask2Former is similar in design to the previous MaskFormer \cite{cheng2021per} architecture. The main components are the backbone feature extractor, pixel decoder, and the Transformer decoder (Figure \ref{fig:mask2former}). The backbone could be either a CNN-based model or a Transformer based model. As the pixel decoder, they have used a more advanced multi-scale deformable attention Transformer (MSDeformAttn) \cite{zhu2020deformable} in contrast to the feature pyramid network \cite{lin2017feature} used in MaskFormer \cite{cheng2021per}. Masked attention has been used to enhance the effectiveness of the Transformer decoder. 

Despite being a universal architecture for segmentation, Mask2Former still needs to be trained separately for each of the specific tasks. This is a common limitation of the universal architectures for segmentation tasks. Mask2Former has achieved new SOTA performance on all three segmentation tasks (panoptic, instance, semantic) in popular datasets such as COCO and ADE20K and Cityscapes. The semantic segmentation results are compared for ADE20K and Cityscapes datasets in Table \ref{table:benchmark-results}. 

\FloatBarrier
\setlength{\tabcolsep}{0.4em}
\begin{table*}[h!]
  \centering
  \resizebox{\columnwidth}{!}{%
    \begin{tabular}{ l l l c c c c }
     \cline{5-7}
     \multicolumn{1}{c}{} & \multicolumn{1}{c}{} & \multicolumn{1}{c}{} & \multicolumn{1}{c}{} & \multicolumn{3}{c}{Datasets} \\
     \hline
     Model & Variant & Backbone & \#Params (M) & ADE20K & Cityscapes & PASCAL-Context\\
     \hline  
     \multirow{8}{*}{SETR \cite{zheng2021rethinking}}  
                            & SETR-\textit{Na\"ive}(16,160k)$^\rho$  & ViT-L$^\ddag$ \cite{Dosovitskiy2020} & 305.67 & 48.06 / 48.80 & - & -\\
                            & SETR-\textit{PUP}(16,160k)      & ViT-L$^\ddag$ & 318.31 & 48.58 / 50.09 & - & -\\
                            & SETR-\textit{MLA}(16,160k)      & ViT-L$^\ddag$ & 310.57 & \textbf{48.64 / 50.28} & - & -\\
                            & SETR-\textit{PUP}(16,40k)       & ViT-L$^\ddag$ & 318.31 & - & 78.39 / 81.57 & -\\
                            & SETR-\textit{PUP}(16,80k)       & ViT-L$^\ddag$ & 318.31 & - & \textbf{79.34 / 82.15} & -\\

                            & SETR-\textit{Na\"ive}(16,80k)   & ViT-L$^\ddag$ & 305.67 & - & - & 52.89 / 53.61\\
                            & SETR-\textit{PUP}(16,80k)       & ViT-L$^\ddag$ & 318.31 & - & - & 54.40 / 55.27\\
                            & SETR-\textit{MLA}(16,80k)       & ViT-L$^\ddag$ & 310.57 & - & - & \textbf{54.87 / 55.83}\\
    \hline                        
    \multirow{4}{*}{Swin $^\aleph$ \cite{liu2021swin}}    & {}  & Swin-T          & 60  & 46.1 & - & -\\
                                                & {}  & Swin-S          & 81  & 49.3 & - & -\\
                                                & {}  & Swin-B$^\ddag$  & 121 & 51.6 & - & -\\
                                                & {}  & Swin-L$^\ddag$  & 234 & \textbf{53.5} & - & -\\
     \hline
     \multirow{4}{*}{Segmenter $^\S$ \cite{strudel2021segmenter}} 
                                & Seg-B         & DeiT-B$^\dag$ \cite{touvron2021training} & 86 & 48.05 & 80.5 & 53.9\\
                                & Seg-B/Mask    & DeiT-B$^\dag$ & 86 & 50.08 & 80.6 & 55.0\\
                                & Seg-L         & ViT-L$^\ddag$ & 307 & 52.25 & 80.7 & 56.5\\
                                & Seg-L/Mask    & ViT-L$^\ddag$  & 307 & \textbf{53.63} & \textbf{81.3} & \textbf{59.0}\\
     \hline
     \multirow{6}{*}{SegFormer \cite{xie2021segformer}} 
                                & {}        & MiT-B0$^\dag$ & 3.4 & 37.4 / 38.0 & 76.2 / 78.1 & -\\
                                & {}        & MiT-B1$^\dag$ & 13.1 & 42.2 / 43.1 & 78.5 / 80.0 & -\\
                                & {}        & MiT-B2$^\dag$ & 24.2 & 46.5 / 47.5 & 81.0 / 82.2 & -\\
                                & {}        & MiT-B3$^\dag$ & 44.0 & 49.4 / 50.0 & 81.7 / 83.3 & -\\
                                & {}        & MiT-B4$^\dag$ & 60.8 & 50.3 / 51.1 & 82.3 / 83.9 & -\\
                                & {}        & MiT-B5$^\dag$ & 81.4 & \textbf{51.0 / 51.8} & \textbf{82.4} / \textbf{84.0} & -\\
    \hline
     \multirow{11}{*}{PVT $^\aleph$}  & {}                            & PVT-Tiny$^\ddag$ & 17.0 & 35.7 & - & -\\
                            & {}                            & PVT-Small$^\ddag$ & 28.2 & 39.8 & - & -\\
                            & {PVT v1 \cite{wang2021pyramid}}& PVT-Medium$^\ddag$  & 48.0 & 41.6 & - & -\\
                            & {}                            & PVT-Large$^\ddag$  & 65.1 & 42.1 & - & -\\
                            & {}                            & PVT-Large$^\ddag$*  & 65.1 & \textbf{44.8} & - & -\\ \cline{2-7}
                            
                            & {}                            & PVT v2-B0$^\ddag$ & 7.6 & 37.2 & - & -\\
                            & {}                            & PVT v2-B1$^\ddag$ & 17.8 & 42.5 & - & -\\
                            & {PVT v2 \cite{wang2022pvt}}   & PVT v2-B2$^\ddag$ & 29.1 & 45.2 & - & -\\
                            & {}                            & PVT v2-B3$^\ddag$ & 49.0 & 47.3 & - & -\\
                            & {}                            & PVT v2-B4$^\ddag$ & 66.3 & 47.9 & - & -\\
                            & {}                            & PVT v2-B5$^\ddag$ & 85.7 & \textbf{48.7} & - & -\\
     \hline
     \multirow{6}{*}{Twins \cite{chu2021twins}} 
                                & {}            & Twins-PCPVT-S$^\dag$ & 54.6 & 46.2 / 47.5 & - & -\\
                                & {Twins-PCPVT} & Twins-PCPVT-B$^\dag$ & 74.3 & 47.1 / 48.4 & - & -\\
                                & {}            & Twins-PCPVT-L$^\dag$ & 91.5 & \textbf{48.6 / 49.8} & - & -\\
                                \cline{2-7}
                                & {}            & Twins-SVT-S$^\dag$ & 54.4 & 46.2 / 47.1 & - & -\\
                                & {Twins-SVT}   & Twins-SVT-B$^\dag$ & 88.5 & 47.7 / 48.9 & - & -\\
                                & {}            & Twins-SVT-L$^\dag$ & 133 & \textbf{48.8 / 50.2} & - & -\\
    \hline
     \multirow{2}{*}{DPT $^\S$ \cite{ranftl2021vision}} & DPT-Hybrid              & ViT-Hybrid$^\ddag$    & 123 & \textbf{49.02} & - & \cellcolor{Gray}\textbf{60.46}\\
                          & DPT-Large            & ViT-L$^\ddag$    & 343   & 47.63 & - & -\\
    \hline
   \multirow{7}{*}{HRFormer \cite{yuan2021hrformer}} 
                            & OCRNet(7,150k)$^\rho$    & HRFormer-S & 13.5 & 44.0 / 45.1 & - & -\\
                                & OCRNet(7,150k)       & HRFormer-B & 50.3 & \textbf{46.3 / 47.6}  & - & -\\
                                & OCRNet(7,80k)        & HRFormer-S & 13.5 & - & 80.0 / 81.0 & -\\
                                & OCRNet(7,80k)        & HRFormer-B & 50.3 & - & 81.4 / 82.0 & -\\
                                & OCRNet(15,80k)       & HRFormer-B & 50.3 & - & \textbf{81.9 / 82.6} & \textbf{57.6 / 58.5}\\
                                & OCRNet(7,60k)        & HRFormer-B & 50.3 & - & - & 56.3 / 57.1\\
                                & OCRNet(7,60k)        & HRFormer-S & 13.5 & - & - & 53.8 / 54.6\\

    \hline
     \multirow{5}{*}{Mask2Former \cite{cheng2022masked}} 
                                & {}        & Swin-T                & -   & 47.7 / 49.6 & - & -\\
                                & {}        & Swin-L$^\ddag$        & 216 & 56.1 / 57.3 & - & -\\
                                & {}        & Swin-L-FaPN$^\ddag$   & -   & \cellcolor{Gray}\textbf{56.4 / 57.7} & - & -\\
                                & {}        & Swin-L$^\ddag$        & 216 & - & 83.3 / 84.3 & -\\
                                & {}        & Swin-B$^\ddag$        & -   & - & \cellcolor{Gray}\textbf{83.3 / 84.5} & -\\

    \hline
    \end{tabular}%
    }
    \caption{\textbf{Comparison of the ViT models specialized for the task of semantic segmentation according to mIoU (\%) using different benchmark datasets. The best-performing variant of each model for a given dataset is highlighted. Overall top performing model variant for each dataset is shaded in gray.} "SS / MS" contains both single-scale and multi-scale inferences.   
    "$\aleph$" - Single-scale inference only,
    "$\S$" - Multi-scale inference only,
    "$\rho$" - (patch size, iterations),
    "\dag" - pre-trained on ImageNet-1K, 
    "\ddag" - pre-trained on ImageNet-21K,  
    "$*$" - 320K training iterations and multi-scale flip testing}
    \label{table:benchmark-results}
\end{table*}
\FloatBarrier

\section{Discussion} \label{Discussion}

In this survey, we discussed how ViTs became a powerful alternative to classical CNNs in various computer vision applications, their strengths as well as limitations, and how ViT contributed to the semantic segmentation of images with their usage across different domains such as remote sensing, medical and video processing. Even though we included some of the CNN architectures widely used in prior mentioned domains to provide a comparison between the ViT and CNNs,  an in-depth discussion about CNN architectures is beyond the scope of this paper. We have summarized the different statistics regarding popular datasets used for semantic segmentation tasks and the results of different ViT architectures used for semantic segmentation to give a clear and high-level overview for the reader around the region of semantic segmentation.

\section{Conclusions and Future Directions} \label{Conclusions and Future Directions}

Unlike mature convolutional neural networks, ViTs are still in the early stage of development. Nevertheless, we observed how powerful and competitive they are with their CNN counterparts. ViTs are progressing towards excellence and it is expected that they will replace traditional CNN-based methods widely used in the deep learning domain in the near future. Different variants of ViTs can be used for experiments with domains such as big data analytics that require a vast amount of data for processing. Exploring research areas with less adaptation to ViT usage can create more efficient, performance-increased outcomes for current implementation methods.

Even though ViTs have proven successful, they can be challenging to experiment with due to their high computational demand. Thus improvements to the ViT architecture are needed to make it lightweight and more efficient. This will inspire the community to open new pathways using ViTs. 

We believe there is a plethora of new research areas that ViT, along with semantic segmentation can be applied to solve real-world problems.



 \bibliographystyle{elsarticle-num} 
 \bibliography{cas-refs}





\end{document}